\documentclass[letterpaper, 10 pt, conference]{ieeeconf}  

\IEEEoverridecommandlockouts                              

\usepackage{times}
\pdfoutput=1 

\usepackage{multicol}
\usepackage{multirow}

\usepackage[utf8]{inputenc} 
\usepackage[T1]{fontenc}    
\usepackage{url}            
\usepackage{booktabs}       
\usepackage{amsfonts}       
\usepackage{nicefrac}       
\usepackage{microtype}      

\usepackage[noend]{algpseudocode}
\usepackage{algorithm}

\usepackage{wrapfig}
\usepackage{caption}

\newcommand{\senet}[0]{\textbf{SE3-Net}}
\newcommand{\senets}[0]{\textbf{SE3-Nets}}
\renewcommand{\senet}[0]{\textsc{SE3-Net}}
\renewcommand{\senets}[0]{\textsc{SE3-Nets}}

\newcommand{\seposenet}[0]{\textbf{SE3-Pose-Net}}
\newcommand{\seposenets}[0]{\textbf{SE3-Pose-Nets}}
\renewcommand{\seposenet}[0]{\textsc{SE3-Pose-Net}}
\renewcommand{\seposenets}[0]{\textsc{SE3-Pose-Nets}}

\usepackage[disable,textsize=tiny]{todonotes}
\usepackage[caption=false]{subfig}

\usepackage{amsmath}
\usepackage{mathtools}
\usepackage{cancel}
\usepackage{amssymb}
\usepackage{color}
\usepackage{makecell}

\DeclareMathOperator*{\SE3}{\mathbb{SE}(3)}
\DeclareMathOperator*{\SO3}{\mathbb{SO}(3)}
\DeclareMathOperator*{\R}{\mathbb{R}}

\newcommand{\mP}{\mathbf{p}}
\newcommand{\mX}{\mathbf{x}}
\newcommand{\mM}{\mathbf{m}}
\newcommand{\mU}{\mathbf{u}}
\newcommand{\mDP}{\Delta{\mP}}

\usepackage[bookmarks=true]{hyperref}
\hypersetup{
    colorlinks=true,
    linkcolor=blue,
    filecolor=magenta,      
    urlcolor=cyan,
}
\usepackage{hyperref}
\usepackage{cite}

\title{SE3-Pose-Nets: Structured Deep Dynamics Models for Visuomotor Planning and Control}
\author{Arunkumar Byravan, Felix Leeb, Franziska Meier and Dieter Fox \\
Department of Computer Science \& Engineering \\
University of Washington, Seattle
}

\begin{document}

\maketitle
\thispagestyle{empty}
\pagestyle{empty}

\begin{abstract}
  In this work, we present an approach to deep visuomotor control using structured deep dynamics models. Our deep dynamics model, a variant of SE3-Nets, learns a low-dimensional pose embedding for visuomotor control via an encoder-decoder structure. Unlike prior work, our dynamics model is structured: given an input scene, our network explicitly learns to segment salient parts and predict their pose-embedding along with their motion modeled as a change in the pose space due to the applied actions. We train our model using a pair of point clouds separated by an action and show that given supervision only in the form of point-wise data associations between the frames our network is able to learn a meaningful segmentation of the scene along with consistent poses. We further show that our model can be used for closed-loop control directly in the learned low-dimensional pose space, where the actions are computed by minimizing error in the pose space using gradient-based methods, similar to traditional model-based control. We present results on controlling a Baxter robot from raw depth data in simulation and in the real world and compare against two baseline deep networks. Our method runs in real-time, achieves good prediction of scene dynamics and outperforms the baseline methods on multiple control runs. Video results can be found at: \url{https://rse-lab.cs.washington.edu/se3-structured-deep-ctrl/}
\end{abstract}

\section{Introduction}
Imagine we are receiving observations of a scene from a camera and we would like to control our robot to reach a target scene. Traditional approaches to visual servoing \cite{hutchinson1996tutorial} decompose this problem into two parts: data-associating the current scene to the target (usually through the use of features) and modeling the effect of applied actions to changes to the scene, combining these in a tight loop to servo to the target. 
Recent work on deep learning has looked at learning similar predictive models directly in the space of observations, relating changes in pixels or 3D points directly to the applied actions \cite{boots2014learning,finn2016unsupervised,byravan2017se3}. Given a target scene, we can use this predictive model to generate suitable controls to visually servo to the target using model-predictive control \cite{finn2017deep}. Unfortunately, for this pipeline to work, we need an external system (such as \cite{schmidt2014dart, anderson2016jump}) capable of providing long range data associations to measure progress.  

As we showed in prior work~\cite{byravan2017se3}, instead of reasoning about raw pixels, we can predict scene dynamics by decomposing the scene into objects and predicting object dynamics instead. While this significantly improves prediction results, it still does not provide a clear solution to the data-association problem that we encounter during control - we still lack the capability to explicitly associate objects/parts across scenes. We observe three key points: 1) We can data-associate across scenes by learning to predict the poses of detected objects/parts in the scene (the pose implicitly provides tracking), 2) We can model the dynamics of an object directly in the predicted low-dimensional pose space, and 3) We can predict scene dynamics by combining the dynamics predictions of each detected part.

\begin{figure}
    \centering
    \includegraphics[width=0.23\textwidth]{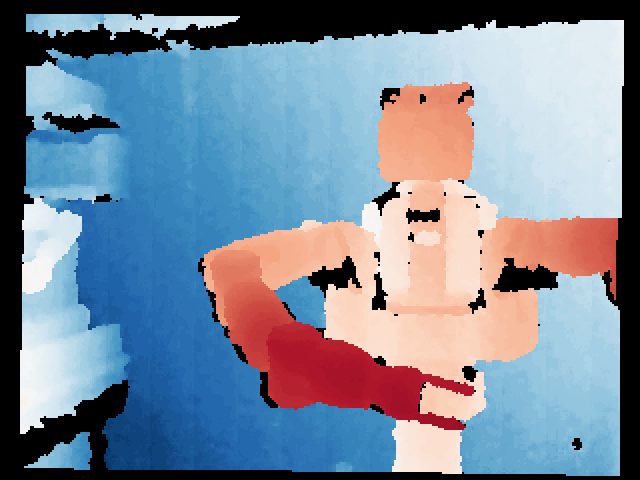}
    \includegraphics[width=0.23\textwidth]{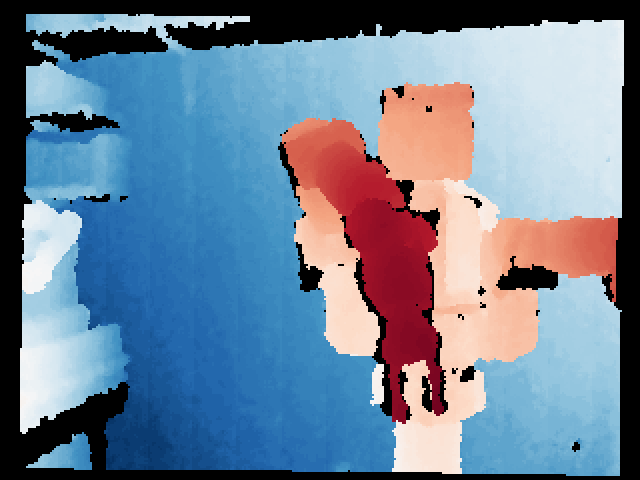}
    \caption{An example scenario showing the initial (left) and target point cloud (right). \seposenets\ can be used to control the robot to reach the target state based only on raw depth data. Depth images colorized for display purposes only.}
    \label{fig:illustration}
    \vspace{-4ex}
\end{figure}

We combine these ideas in this work to propose \seposenets, a deep network architecture for efficient visuomotor control that jointly learns to data-associate across long term sequences. We make the following contributions:
\begin{itemize}
    \item We show how it is possible to learn predictive models that detect parts of the scene and jointly learn a consistent pose space for these parts with minimal supervision.
    \item We demonstrate how a deep predictive model can be used for reactive visuo-motor control using simple gradient backpropagation and a more sophisticated Gauss-Newton optimization, reminiscent of approaches in inverse kinematics \cite{buss2004introduction}.
    \item We present results on real-time reactive control of a Baxter arm using raw depth images and velocity control, both in simulation and on real data.
\end{itemize}
Fig~\ref{fig:illustration} shows an example scenario where our proposed method can be applied to control the robot to reach the target state (right) from the initial state (left). 



\section{Related work}
\textbf{Modeling scenes and dynamics: } Our work builds on top of prior work on learning structured models of scene dynamics \cite{byravan2017se3}. Unlike \senets\, we now explicitly model data associations through a low-dimensional pose embedding that we train to be consistent across long sequences. Similar to Boots et al. \cite{boots2014learning}, our model learns to predict point clouds based on applied actions, but through a more structured intermediate representation that reasons about objects and their motions. Unlike Finn et al. \cite{finn2016unsupervised}, we operate on depth data and reason about motion in 3D using masks and $\SE3$ transforms while training our networks in a supervised fashion given point-wise data associations across pairs of frames.

\textbf{Visuomotor control: } Recently, there has been a lot of work on visuomotor control, primarily through the use of deep networks \cite{watter2015embed, wahlstrom2015pixels, levine2016end, agrawal2016learning, finn2017deep, jonschkowski2017pves}. These methods either directly regress to controls from visual data \cite{levine2016end, wahlstrom2015pixels}, generate controls by planning on learned forward dynamics models \cite{finn2017deep, watter2015embed}, through inverse dynamics models \cite{agrawal2016learning} or by reinforcement learning  \cite{jonschkowski2017pves}. Similar to some of these methods, we generates controls by planning with a learned dynamics model, albeit in a learned low-dimensional latent space.
Specifically, work by Finn et al. \cite{finn2017deep} is closely related, but differs in two main ways: unlike their approach which controls in the observation space through sampled actions (at $\approx$ 5Hz), our controller runs gradient based optimization on a learned low-dimensional pose embedding in real-time (> 30 Hz). Also, their approach requires an external tracker to measure progress while we explicitly learn to data associate across large motions. \\
Our work borrows several ideas from prior work by Watter et al. \cite{watter2015embed} which learns a latent low-dimensional embedding for fast reactive control from pairs of images related by an action. Unlike their work though, we use a structured latent representation (object poses), predict object masks and use a physically grounded 3D loss that only models change in observations as opposed to a restrictive image reconstruction loss. Last, our losses are physically motivated similar to those proposed for training position-velocity encoders \cite{jonschkowski2017pves}, but our learned pose embedding is significantly more structured and we train our networks end-to-end directly for control.



\textbf{Data association: } Related work in the computer vision literature has looked at tackling the data association problem, primarily by matching visual descriptors, either hand-tuned \cite{lowe2004distinctive}, or more recently, learned using deep networks \cite{schmidt2017self, wang2015unsupervised}. In prior work, Schmidt et al. \cite{schmidt2017self} learn robust visual descriptors for long-range associations using correspondences over short training sequences. Unlike this work, we only use correspondences between pairs of frames to learn a consistent pose space that lets us data associate across long sequences.


\textbf{Visual servoing: } Finally, there have been multiple approaches to visual servoing over the years \cite{hutchinson1996tutorial, espiau1993new, ChaumetteSB16}, including some newer methods that use deep learned features and reinforcement learning \cite{lee2017learning}. While these methods depend on an external system for data association or on pre-specified features, our system is trained end-to-end and can control directly from raw depth data.

\section{\seposenets}

\begin{figure*}
  \begin{center}
    \centering
    \vspace{1mm}
    \includegraphics[width=0.8\textwidth]{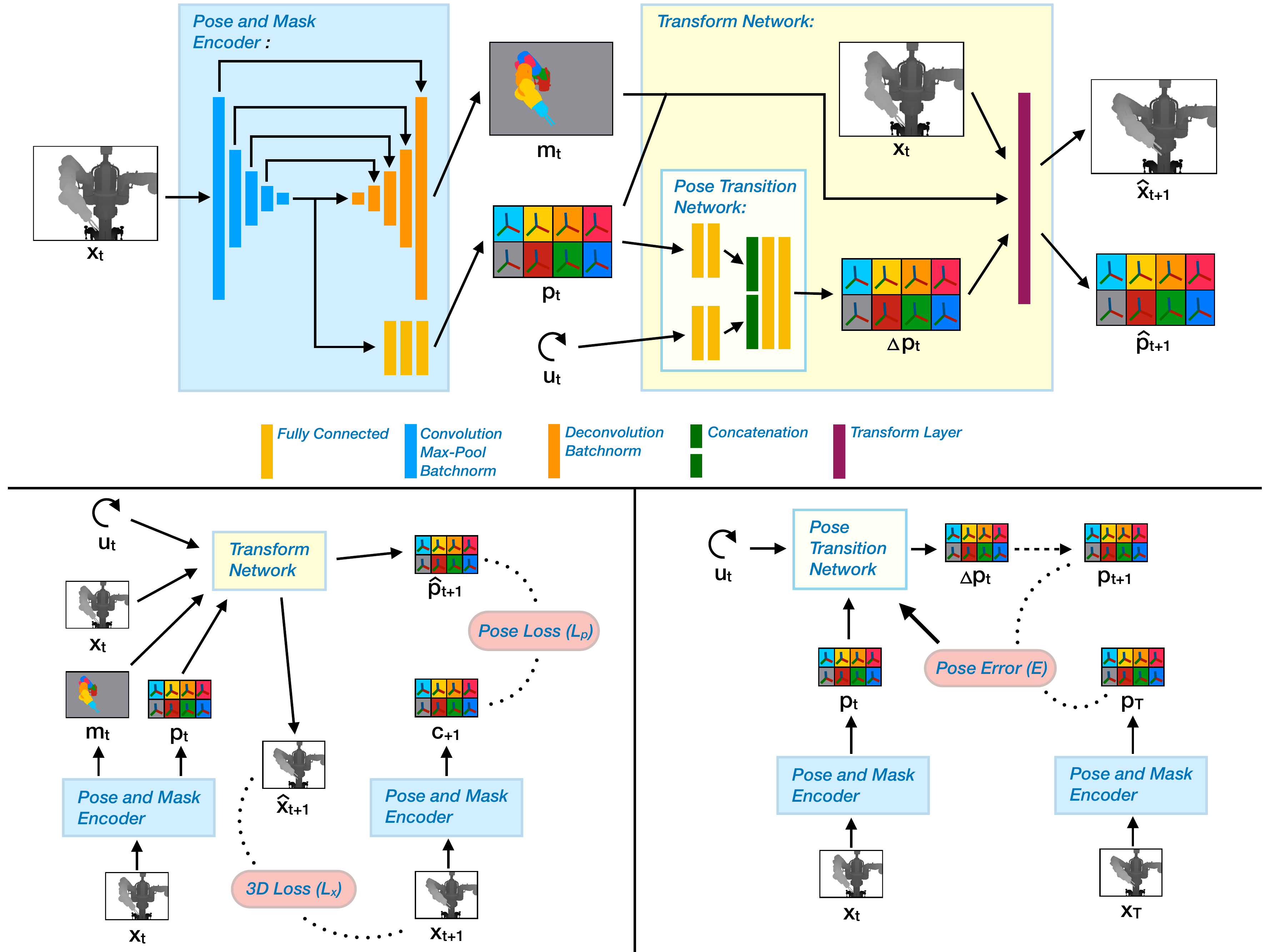}
    	\vspace{-1mm}
    	\vspace*{-1ex}
    \caption{\small{\textbf{Top:} \seposenet\ architecture consisting of three components: the \textbf{encoder} ($h_{enc}$, shown in blue) that predicts dense segmentation masks ($\mM$) and 6D poses ($\mP$), a \textbf{pose transition net} ($h_{trans}$) that models the change in the pose space ($\mDP$) as an effect of the applied action ($\mU$) and the \textbf{transform layer} that applies these pose changes to the current point cloud to generate a predicted point cloud ($\hat{\mX}$). \textbf{Bottom Left:} Graph showing the procedure for training the \seposenet\ along with two loss functions: a 3D loss on the predicted point cloud ($L_x$) and a pose consistency loss ($L_p$) relating the "next" poses predicted by the transform network ($\hat{\mP}_{t+1}$) and the encoder ($\mP_{t+1}$). \textbf{Bottom Right: } Control using the \seposenet. Given a target point cloud ($\mX_T$) encoded as poses ($\mP_T$) through the learned encoder, we use the learned transition model ($h_{trans}$) to plan a sequence of actions $\mU_0, \mU_1,...,\mU_T$ by minimizing error (E) directly in the pose space from an initial point cloud $\mP_0$.}}\vspace*{-5ex}
    \label{fig:network}
  \end{center}
\end{figure*}

Our deep dynamics model \seposenets\ decomposes the problem of modeling scene dynamics into three sub-problems: a) modeling scene structure by identifying parts of the scene that move distinctly and by encoding their latent state as a 6D pose, b) modeling the dynamics of individual parts under the effect of the applied actions as a change in the latent pose space (parameterized as an $\SE3$ transform), and finally c) combining these local pose changes to model the dynamics of the entire scene. Each sub-problem is modeled by a separate component of the \seposenet:
\begin{itemize}
    \item \textbf{Modeling scene structure:} An \textit{\textbf{encoder}} ($h_{enc}$) that decomposes the input point cloud ($\mX$\footnote{Bold fonts denote collections of items}) into a set of $K$ rigid parts, predicting per part a 6D pose ($p^k$, $k=1 \dots K$) and a dense segmentation mask ($m^k$) that highlights points belonging to that part 
    \item \textbf{Modeling part dynamics:} A \textit{\textbf{pose transition}} network ($h_{trans}$) that models dynamics in the pose space, taking in the current poses ($\mP_t$) and action ($\mU_t$) to predict the change in poses ($\mDP_t$)
    \item \textbf{Predicting scene dynamics:} A \textit{\textbf{transform}} layer ($h_{tfm}$) that predicts the next point cloud ($\hat{\mX}_{t+1}$) given the current point cloud ($\mX_t$), predicted object masks ($\mM_t$) and the predicted pose deltas ($\mDP_t$) by explicitly applying 3D rigid body $\SE3$ transforms on the input point cloud.
\end{itemize}
Fig.~\ref{fig:network} shows the network architecture of the \seposenet. Next, we present the details of the three sub-components and outline a training procedure for training the \seposenet\ end-to-end with minimal supervision.

\subsection{Modeling scene structure}
Given a 3D point cloud $\mX$ from a depth sensor (represented as a 3-channel image, 3 x H x W), the \textbf{encoder} (blue block in Fig.~\ref{fig:network}) segments the scene into distinctly moving parts ($\mM$) and predicts a 6D pose per segmented part ($\mP$).
\begin{align}
(\mP, \mM) = h_{enc}(\mX)
\end{align}
The encoder has three parts, the first is a convolutional network that generates a latent representation of the input point cloud ($\mX$). This network has five convolutional layers, each followed by a max pooling layer. The latent representation is further used as input for the mask and pose predictions.

\textbf{Object masks: }
We use a de-convolutional network to predict a dense pixel-wise segmentation of the scene into it's constituent parts ($\mM$). Similar to prior work \cite{byravan2017se3}, we use a fully-convolutional architecture with five de-convolutional layers and a skip-add architecture to improve the sharpness of the predicted segmentation. The masks predicted by this network are at full resolution with $K$ channels (K x H x W), where $K$ is a pre-specified hyper-parameter that is greater than or equal to the number of moving parts in the scene (including background). The predicted segmentation mask learns to attend to parts of the scene that move together, representing areas of the scene that can move independently as different parts.
As in prior work \cite{byravan2017se3}, we formalize mask prediction as a soft-classification problem where the network outputs a $k$-length probability distribution which we sharpen to push towards a binary segmentation mask.

\textbf{Object poses: }
Given the encoded latent representation, we use a three layer fully-connected network to predict the 6D pose $p^k$ of each of the $K$ segmented parts. We represent each pose by 6 numbers: a 3D position ($y \in \R^3$) and an orientation ($R \in \SO3$), represented as a 3-parameter axis-angle vector. As we show later, our pose network learns to predict consistent poses which can be used to data-associate observations over long sequences of motions.

At a high level, the \textbf{encoder} implicitly learns the structure of observed scenes by persistently identifying parts and predicting a consistent pose for each part across multiple scenes.

\subsection{Modeling part dynamics}
Once we have identified the constituent parts of the scene and their poses, we can reason about the effect of applied actions on these parts. We model this notion of "part dynamics" through a fully-connected \textbf{pose transition} network that takes the predicted poses from the encoder ($\mP$) and applied actions ($\mU$) as input to predict the change in pose ($\mDP$) for all $K$ segmented parts:
\vspace{-1ex}
\begin{align}
    \mDP = h_{\text{trans}}(\mP, \mU)
\end{align}
where $\mDP = [\textbf{R}, \textbf{T}]$ is represented as an $\SE3$ transform per part, with a rotation $\text{R}^k \in \SO3$ (parameterized as an axis-angle transform) and a translation vector $\text{T}^k \in \R^3$. The transition network first applies two fully connected layers to both inputs, concatenates their outputs followed by two final fully-connected layers to predict the pose-deltas. As we show later in Sec.~\ref{sec:ctrl} we rely on good predictions of pose-deltas through the pose-transition network for efficient control.



\subsection{Predicting scene dynamics}
Finally, given the predicted scene segmentation ($\mM_t$) and the change in poses ($\mDP_t$), we can model the dynamics of the input scene ($\mX_t$) under the effect of the applied action ($\mU_t$). We do this through the \textbf{Transform} layer ($h_{tfm}$) which applies the predicted rigid rotations ($\mathbf{R}_t$) and translations ($\mathbf{T}_t$) to the input point cloud, weighted by the predicted mask probabilities ($\mM_t$). We predict the transformed point cloud ($\hat{\mX}_{t+1}$) as: 
\vspace{-4mm}
\begin{align}
\hat{x}^j_{t+1} = \sum_{k=1}^{K} m_{t}^{kj} \left( R^k_t \; x^j_t + T^k_t \right)  
\vspace{-4mm}
\label{eq:finaltfm} 
\end{align}
where $\hat{x}^j_{t+1}$ is the 3D output point corresponding to input point $x^j_t$. In effect, we apply the $k$th rotation and translation ($\Delta{p}^k = [R^k, T^k])$ to all points $x^j$ that belong to the corresponding object as indicated by the $k$th mask channel $m^k$ (assuming that the mask is binary after weight sharpening) to predict the transformed points $\hat{x}^j$ belonging to that object. Repeating this for all objects gives us the transformed output point cloud ($\hat{\mX}$). Note that this part has no trainable parameters. For more details, please refer to prior work \cite{byravan2017se3}.

\subsection{Training}
We now outline a procedure to train the \seposenet\ end-to-end, using supervision in the form of point-wise data associations across a pair of point clouds ($\mX_t$, $\mX_{t+1}$), related by an action ($\mU_t$) i.e. for each input point ($x_t^i$), we know it's corresponding point in the next frame if it is visible ($x_{t+1}^i$). No other supervision is given for learning the masks, poses, and the change in poses. Fig.~\ref{fig:network} (bottom-left) shows a schematic of this procedure.
Given two point clouds $\mX_t, \mX_{t+1}$, we use the \textbf{encoder} to predict the corresponding masks and poses:
\vspace{-3ex}
\begin{align}
\mP_t, \mM_t &= h_{enc}(\mX_t) \nonumber \\ 
\mP_{t+1}, \mM_{t+1} &= h_{enc}(\mX_{t+1})
\end{align}
Next, the predicted pose ($\mP_t$) and control ($\mU_t$) at $t$ are used as input to the \textbf{pose transition} network to predict the change in pose from $t$ to $t+1$:
\vspace{-1ex}
\begin{align}
\mDP_{t} = h_{trans}(\mP_t, \mU_t) 
\end{align}
Finally, we use the \textbf{transform} layer (\ref{eq:finaltfm}) to predict the next point cloud:
\vspace{-2ex}
\begin{align}
\hat{\mX}_{t+1} &= h_{tfm}(\mX_t, \mM_t, \mDP_t)
\end{align}
The predicted mask ($\mM_{t+1}$) at time $t+1$ is discarded.
We use two losses to train the entire pipeline end to end:
\begin{itemize}
    \item A 3D loss ($L_x$) that penalizes the error between the predicted point cloud ($\hat{\mX}_{t+1}$) and the data associated target point cloud ($\tilde{\mX}_{t+1}$). We use a normalized version of the mean-squared error (MSE) that measures the negative log-likelihood under a Gaussian centered around the target with a standard deviation dependent on the target magnitude:
    \vspace{-2ex}
    \begin{align} \label{eq:3dloss}
        L_x = \frac{1}{N} \sum_{i=1}^{HW} \dfrac{(\hat{x}_{t+1}^i - \tilde{x}_{t+1}^i)^2}{\alpha {\tilde{f}}^i + \beta}
    \end{align}
    where ($\tilde{f}^i = \tilde{x}_{t+1}^i - x_t^i$) denotes the ground truth motion for point $i$ relative to the input point cloud $\mX_t$, $HW$ is the number of points in the point cloud, $N$ is the number of points that actually move between $t$ and $t+1$ and $\alpha$ \& $\beta$ are hyper-parameters ($\alpha = 0.5, \beta = 1e-3$ in all our experiments). This loss is aimed to tackle two main issues with a standard MSE loss: a) By normalizing the loss by a separate scalar per dimension ($\tilde{f}^i$) that depends on the target magnitude we make the loss scale invariant allowing us to treat equally parts that move less (such as the end-effector when only the wrist rotates) as those that have large motion (eg. the elbow), and b) By dividing the total error by the number of points ($N$) that move in the scene, we treat scenes where very few points move equally as those where large parts move. 
    \item A pose consistency loss ($L_p$) that encourages consistency between the poses predicted by the encoder ($\mP_t, \mP_{t+1}$) and the change in pose predicted by the pose transition network ($\mDP_t$):
    \vspace{-2ex}
    \begin{align} \label{eq:consisloss}
        \hat{\mP}_{t+1} = \mP_t \oplus \mDP_t \nonumber \\
        L_p = \frac{1}{I} \sum_{i=1}^{I} (\hat{p}^i_{t+1} - p^i_{t+1})^2
    \end{align}
    where $\oplus$ refers to composition in $\SE3$ pose space, $\hat{\mP}_{t+1}$ is the expected pose at $t+1$ from composing the current pose ($\mP_t$) and the predicted pose change from the transition model ($\mDP_t$) and $I$ is the cardinality of $\mP_t$. In essence, this loss constrains the encoder to predict poses that are consistent with the pose-deltas predicted by the transition model. This loss encourages global consistency in the pose space by enforcing local consistency over pairs of frames and is crucial for learning a pose space that is consistent across long term motions. 
\end{itemize}
The total loss for training ($L$) is a sum of the two losses: $L = L_x + \gamma L_p$, where $\gamma$ controls the relative strengths of the two losses. We set $\gamma = 10$ in all our experiments. A key point to note is that we do not provide any explicit supervision to learn the pose space. While the consistency loss ensures that the poses are more or less globally consistent, it does not anchor them to a specific 3D position or orientation. As such, the poses learned by the network need not correspond directly to the canonical 6D pose of the parts - the predicted part position $y^k$ does not need to correspond to its center and the orientation need not be aligned to the part's principal axes. Providing more constraints to regularize and physically ground the pose space is an interesting area for future work.

\section{Closed-Loop Visuomotor Control using \seposenets} \label{sec:ctrl}

We now show how an \seposenet\ can be used for closed-loop visuomotor control to reach a target specified as a target depth image, essentially performing visual servoing~\cite{hutchinson1996tutorial}.  A crucial component of every visual servoing system is to perform data association between the current image and the target image, which can then be used to generate controls that reduce the corresponding offsets. \seposenets\ solve this problem by making use of the learned, low-dimensional latent pose space.  By enforcing frame-to-frame consistency in the pose space through the consistency loss (Eqn.~\ref{eq:consisloss}), the pose space becomes consistent, that is, our encoder network  learns to data-associate observations to unique poses which are consistent under the effect of actions. Importantly, these data associations are generated at the mask, or object level, resulting in an ability akin to object detection in computer vision. Unlike prior work \cite{finn2016deep, byravan2017se3} which is restricted to operate in the observation space of 3D points and requires data associations between current and target points to be provided externally, we can now directly minimize error between the poses $\mP_0$ and $\mP_T$ automatically extracted from the initial and the target depth image to recover the sequence of actions that takes the robot from $\mP_0$ to $\mP_t$. Additionally, unlike prior work \cite{finn2016deep}, we do not need an external tracking system to measure progress toward the goal as our learned \textbf{encoder} implicitly tracks in the pose space.

\vspace{-1ex}
\subsection{Reactive control}
\label{sec:reactivectrl}
Algorithm \ref{alg:reactivectrl} presents a simple algorithm for reactive control using \seposenets\ that efficiently computes a closed-loop sequence of controls that takes the robot from any initial state $\mX_0$ to the specified target $\mX_T$ (the corresponding network structure is given in the lower right panel of Fig.~\ref{fig:network}).  Given a target point cloud, $\mX_T$, the algorithm uses the learned encoder to predict the  poses of the constituent parts $\mP_T = h_{enc}(\mX_T)$. This becomes the target to the controller. 

\begin{algorithm} [t]
\caption{Reactive visuomotor control}
\label{alg:reactivectrl}
\begin{algorithmic}
\State Given: Target point cloud ($\mX_T$) 
\State Given: Pre-trained encoder ($h_{enc}$) and transition model ($h_{trans}$)
\State Given: Maximum control magnitude: $u_{max}$
\State Compute target pose: $\mP_T = h_{enc}(\mX_T)$
\While {not converged}
   \State Receive current observation ($\mX_t$)
   \State Predict current pose: $\mP_t = h_{enc}(\mX_t)$
   \State Initialize control to all zeros: $\mU_t = 0$
   \State Predict change in pose: $\mDP_t = h_{trans}(\mP_t, \mU_t)$
   \State Predict next pose: $\hat{\mP}_{t+1} = \mP_t \oplus \mDP_t$
   \State Compute pose error: $E = \frac{1}{I} \sum_{i=1}^{I} (\hat{p}^i_{t+1} - p^i_T)^2$
   \State Compute gradient of error w.r.t.~control: $g = \dfrac{dE}{dU_t}$
   \State Compute control: $\mU_t = - u_{max} * \dfrac{g}{||g||}$
   \State Execute control $\mU_t$ on the robot
\EndWhile
\end{algorithmic}
\end{algorithm}
\setlength{\textfloatsep}{0pt}

At every time step, the algorithm computes the pose embedding  $\mP_t$ of the current observation $\mX_t$.  
We would like to find controls that move these poses closer to the target poses. To do this, the algorithm makes a prediction through the learned \textbf{pose transition} model using the current poses ($\mP_t$) and an initial guess for the controls (here we use $\mU_t = 0$), resulting in a predicted change in poses ($\mDP_t$) and the corresponding predicted next pose ($\hat{\mP}_{t+1}$)~\footnote{Even when using a zero control initialization, this forward pass through the network is necessary to get the correct gradients for the backward pass.}.  To move these poses towards the targets, we formulate an error function $E$ based on the mean-squared error between these predicted poses and the target poses.  The algorithm then  computes the gradient of this error with respect to the control inputs, which it uses to generate the next controls. We propose two ways of computing the gradient:
\begin{itemize}
    \item \textbf{Backpropagation:} A simple approach to compute this gradient update is to backpropagate the gradients of the pose error $E$ through the pose transition model. 
%
Unlike backpropagation during training, where we compute gradients w.r.t. the network weights, here we fix the weights and compute gradients over the input controls. The resulting control scheme is analogous to the Jacobian Transpose method from inverse kinematics \cite{buss2004introduction}, where backprop provides the gradient of the transition model. 

    \item \textbf{Gauss-Newton:} A better approach is to compute the Gauss-Newton update:
    \vspace*{-2ex}
    \begin{align}
    g = (J^T J + \lambda I)^{-1} J^T * g_P \label{eq:JacPseudoInv}
    \vspace*{-2ex}
    \end{align}
    where $J$ is the Jacobian of the transition model, and $g_P$ is the gradient of the pose error (E). However, instead of computing $g$ via backpropagation, we condition the pose error gradient ($g_P$) based on the Jacobian's pseudo-inverse, where $\lambda$ controls the strength of the conditioning (set to 1e-4 in all our experiments). In practice, this leads to significantly faster convergence  with little to no additional overhead in computation compared to the backpropagation method as the Jacobian can be computed efficiently through finite differencing. We do this by running a single forward propagation with perturbed control inputs (perturbation set to 1e-3) stacked along the batch dimension to take advantage of GPU parallelism. Eqn.~\ref{eq:JacPseudoInv} is also analogous to the Damped Least Squares technique from inverse kinematics~\cite{buss2004introduction}.
\end{itemize}
Finally, the algorithm computes the unit-vector in the direction of the computed update and scales this by a pre-specified control magnitude $u_{max}$ (1 radian in all our experiments) to get the next control $\mU_t$. We execute this control on the robot and repeat in a closed-loop until \textbf{convergence} measured either by reaching a small error in the pose space ($E < \epsilon$) or a maximum number of iterations, whichever comes first.


\section{Evaluation}
\label{sec:evaluation}
We first evaluate \seposenets\ on predicting the dynamics of a scene where a Baxter robot moves its right arm in front of the depth camera, both in simulation and in the real world. We also present results on control performance where the task is to control the joints of the Baxter's right arm to reach a specified target observation.

\begin{figure*}
  \begin{center}
    \centering
    \vspace{1mm}
    \includegraphics[width=0.8\textwidth]{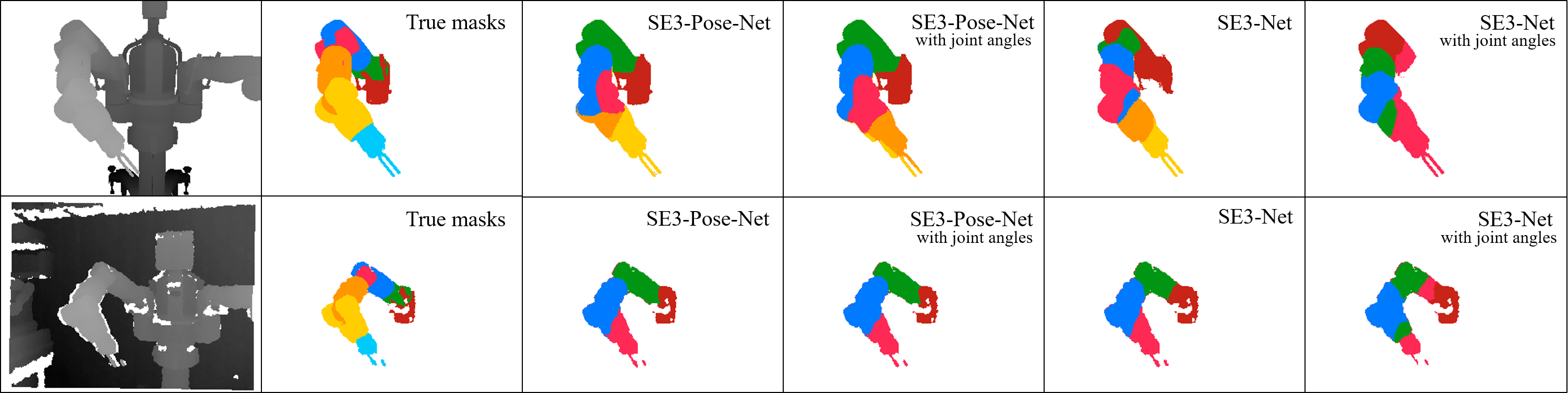}
    	\vspace*{-0.5ex}
    \caption{\small{Masks generated by different networks on simulated (top) and real data (bottom). From left to right: Ground truth depth, ground truth masks, masks predicted by the \seposenet, \seposenet\ with joint angles, \senet\ and \senet\ with joint angles.}}\vspace*{-2.5ex}
    \label{fig:masks}
  \end{center}
\end{figure*} 

\begin{table*}
\centering
\begin{tabular}{| c | c | c | c | c | c | c |}
\hline
Setting          & \seposenets\ & \seposenets\ + Joint Angles & \senets\  & \senets\ + Joint Angles & Flow & Flow + Joint Angles \\
\hline
Simulated 		 & 0.044 & 0.038 & 0.030 & \textbf{0.024} & 0.035 & 0.030  \\
Real             & 0.234 & 0.224 & 0.221 & \textbf{0.212} & 0.228 & 0.218  \\
 \hline
\end{tabular}\vspace*{-1ex}
\caption{\small{Average per-point flow MSE (cm) across tasks and networks, normalized by the number of points $M$ that move in the ground truth data (motion magnitude > 1mm). Our network achieves results slightly worse than the baseline networks on both simulated and real data. However, it is also solving additional tasks necessary for control.}}
\label{tbl:flowerror}\vspace*{-4.5ex}
\end{table*}

\subsection{Task and Data collection}
We first provide details on the task setting in simulation. Our simulator uses OpenGL to render depth images from a  camera pointed towards the robot (see Fig.~\ref{fig:illustration}) and is kinematic with little to no dynamics in the motion and no depth noise. We use this as a test bed to parse the effectiveness of the proposed algorithm and compare it to various baselines. We collected around 8 hours of training data in the simulator where the robot moves all joints on it's right arm. Around half of the examples are whole arm motions where the robot plans a trajectory to reach a target end-effector position sampled randomly in the workspace in front of the robot. The rest of the motions are made of perturbations of individual joints on the robot from various initial configurations sampled to be within the viewpoint of the camera. These additional motions help in de-correlating the kinematic chain dependencies during training, improving performance especially on joints lower down the kinematic chain. Overall, this dataset has around 800,000 training images collected from a single fixed viewpoint. 
Similar to the simulated setting, we collect data from the real robot where the Baxter moves its right arm in front of an ASUS Xtion Pro camera placed around 2.5 meters from the robot. Data associations, ground truth masks, and ground truth flows are determined via the DART tracker~\cite{schmidt2014dart} on the real data.  We collected around 4.5 hours of training data on the real robot, with a 2:1 mix of whole arm motions and single joint motions. As before, the motions were generated through a planner that tries to get the end-effector to randomly sampled targets in the workspace. Unlike the simulated data, the depth data in the real world is quite noisy and there are significant physical and dynamics effects. 
For both the simulated and real world settings, our controls are joint velocities ($\mU$). 

\subsection{Baselines}
We compare the performance of our algorithm against five different baselines:
\begin{itemize}
    \item \textbf{\seposenets\ + Joint Angles: } Our proposed network with the joint angles of the robot given as an additional input to the \textbf{encoder}. We use this network as a strong baseline that uses significant additional information to inform the pose prediction. 
    \item \textbf{\senets: } Prior work from \cite{byravan2017se3} where the network directly predicts masks and change in poses given input point clouds and control. There is no explicit pose space in this network, so we do control in the full point cloud observation space for this network. 
    \item \textbf{\senets\ + Joint Angles: } \senets\ that additionally take in joint angles as inputs.
    \item \textbf{Flow Net: } Baseline flow model from prior work \cite{byravan2017se3}. This network directly regresses to a per-point 3D flow without any explicit $\SE3$ transforms or masks.
    \item \textbf{Flow Net + Joint Angles:} Baseline flow network that additionally takes in joint angles as input.
\end{itemize}
All baseline networks are trained on the same data as the \seposenets\ using the 3D normalized loss ($L_x$). 

\subsection{Training details}
We implemented our networks in PyTorch using the Adam optimizer for training with a learning rate of 1e-4. All our networks used Batch Normalization \cite{ioffe2015batch} and the PReLU non-linearity \cite{he2015delving}. 
We set the maximum number of moving objects $K = 8$ for all our experiments (7 joints + background). We train each network for 100,000 iterations in simulation and 75,000 iterations on the real data, and use the network that achieves the least validation loss across all training iterations for all our results. 

\subsection{Results on modeling scene dynamics}
First, we present results on the prediction task used for training all the networks.  Table \ref{tbl:flowerror} shows the average per-point flow MSE (cm) across all baselines on both simulated and real data. \senets\ achieve the best results on both the simulated and real datasets while the baseline flow network performs slightly worse. Unsurprisingly, networks that have access to the joint angles do better than those which do not, as they have strictly more information that is highly correlated with the sensor data. To initial surprise, the \seposenets\ have the largest prediction errors among all baseline models. However, this makes sense given the following considerations: a) \seposenets\ are trained to explicitly embed the observations in a pose space from which they predict the scene dynamics, rather than using the input point cloud directly. While this provides more structure within the network and is necessary for the control task, it also restricts the prediction to go through an information bottleneck which generally makes the training problem harder. b) \seposenets\ additionally have to optimize for the consistency loss, which enforces constraints that are different from those of the prediction problem evaluated in this experiment. \\
Fig.~\ref{fig:masks} visualizes the masks predicted by \seposenets\ and the baseline \senet\ on an example each from the simulated and real data along with the ground truth masks. Even without any supervision, \seposenets\ and \senets\ learn a detailed segmentation of the arm into multiple salient parts, most of which are consistent with ground truth segments on both the simulated and real data.

\begin{figure*}
\centering
\includegraphics[width=0.32\textwidth]{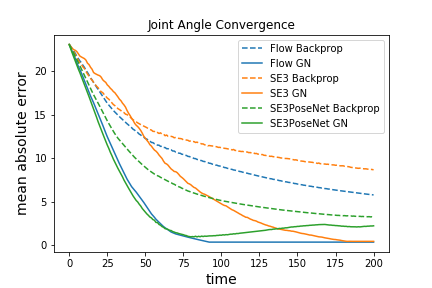}
\includegraphics[width=0.32\textwidth]{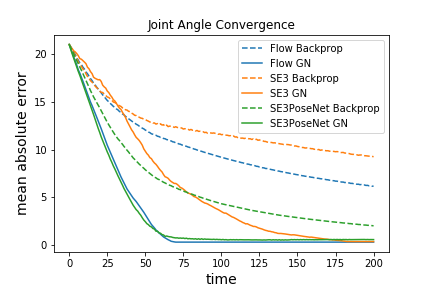}
\includegraphics[width=0.32\textwidth]{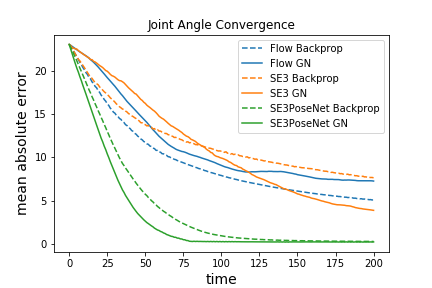}
\vspace{-1mm}
\caption{Convergence of joint angle error in simulated Baxter control tasks. (left): without joint angles, (middle) without joint angles and detected failure case removed (for all methods), (right) with joint angles. \seposenets\ perform as well or better than baseline methods even though baseline models have additional information in the form of ground truth-associations.}
\label{fig:joint_angle_errors_sim}\vspace*{-3.5ex}
\end{figure*}
\begin{figure}
\centering
\includegraphics[width=0.23\textwidth]{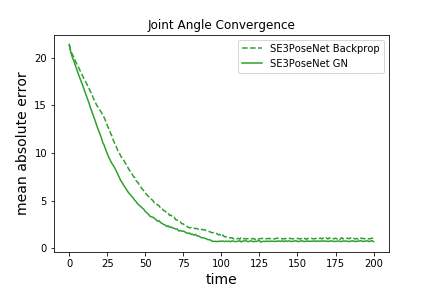}
\includegraphics[width=0.23\textwidth]{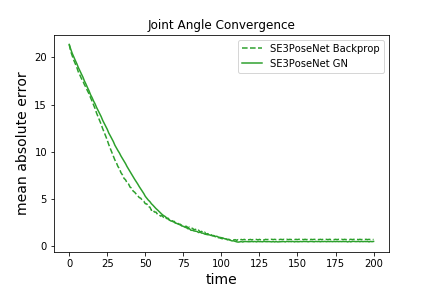}
\caption{Convergence of joint angle error on real Baxter control tasks (left) without joint angles (right) with joint angles (averaged across joint 0,1,2,3).}
\label{fig:joint_angle_errors_real}
\end{figure}

\subsection{Control performance}\label{sec:ctrlresults}
Next, we test the performance of the different networks on controlling the Baxter's right arm to reach a target configuration, specified as a point cloud $\mX_T$. We test both the control algorithms presented in Sec.~\ref{sec:ctrl} using our \seposenet\ model and the baseline models by comparing their performance on a set of 11 distinct servoing tasks (each with an average initial error of \textasciitilde30 degrees per joint). We first detail a few specifics followed by an analysis of the results.

\textbf{Control with baseline models: } While \seposenets\ learn a pose space that can be used for long-term data associations and control, the baseline models operate directly in the space of observations and thus \textbf{require external data associations} in the observation space to be able to do any control at all. For the simulation experiments, we provide these baseline algorithms with ground-truth associations and use the procedure outlined in Alg. \ref{alg:reactivectrl} using the MSE between the predicted point cloud $\hat{\mX}_{t+1}$ and the target $\mX_T$ as the error to be minimized for generating controls. It is important to keep in mind that the baseline models have an advantage over \seposenets\ for the control task as they get strictly more information in the form of ground-truth data associations.

\textbf{Metric and Task specification: } We use the mean absolute error in the joint angles as the metric for measuring control performance. We run all models to convergence (based on the pose error for \seposenets\ and 3D point/flow error for the baseline models) or for a maximum of 200 iterations. Additionally, for \seposenets\, we terminate if the pose error increases for 10 consecutive iterations. We integrate joint velocities forward to generate position commands for the robot both in simulation and the real world.

\textbf{Simulation results: } Fig.~\ref{fig:joint_angle_errors_sim} plots the error in joint angles as a function of the number of control iterations. The plots on the left and middle show results on networks that use only raw depth as input - we control the \textit{first six joints} of the robot using these networks. The right figure shows results for networks that additionally use joint angles as input - we control \textit{all 7 joints} of the robot with these networks. In general, \seposenets\ achieve excellent performance compared to the baseline models, converging quickly to an almost zero error even \textbf{in the absence any external data associations}. The flow model performs comparably to the \seposenets\ while \senets\ converge far slower. We highlight a few key results: 1) For all methods, Gauss-Newton based optimization (GN) leads to faster convergence than Backprop. This is to be expected as Gauss-Newton conditions the gradient based on pseudo-second order information. 2) Baseline models perform worse given joint angles than without. This is due to an issue of credit assignment during gradient computation - the networks learn erroneous causations (when there are only correlations) between the input joint angles and the predicted flows which diminishes the control's contribution to the prediction problem and subsequently affects the gradient. 3) All models struggle to model the motion of the final wrist joint due to increasing correlations along the kinematic chain that result in a small contribution of the joint's own motion to the full movement of the wrist. \seposenets\ can overcome this problem given input joint angles (Fig.~\ref{fig:joint_angle_errors_sim}, right) which provides encouraging proof that adding in the joint state supplements information that is hard to parse directly from the visual state. 4) \senets\ converge slowly due to a lack of good control initializations that are needed to ensure that the network starts off with a meaningful segmentation - given zero controls the \senet\ can choose not to segment the arm at all, and finally 5) Good performance of \seposenets\ indicates that the learned pose space is consistent across large motions and can be used for fast reactive control, albeit not quite as robust as the baseline methods given data associations. \seposenets\ fail to minimize the pose error on one of the tested configurations leading to an increasing error in Fig.~\ref{fig:joint_angle_errors_sim}, left. Our termination check that looks for increasing pose errors does correctly identify this case and we are able to succeed on all the other examples (Fig.~\ref{fig:joint_angle_errors_sim}, middle). We discuss ways to further improve the robustness of our approach in Sec.~\ref{sec:future}.

\textbf{Real robot results: } We further test the control performance using \seposenets\ on a few real world examples. We do not compare to any baselines as they need an explicit external data association system to be feasible. On the real robot, we restrict ourselves to controlling the \textit{first four joints} of the right arm using the \seposenet\ and control the \textit{first six joints} using the model that additionally takes in joint angles as input. Fig.~\ref{fig:joint_angle_errors_real} shows the errors as a function of the iteration count. Both models converge very quickly which indicates that our network is able to control robustly even in the presence of sensor noise and unmodeled dynamics. Surprisingly, there is very little difference between GN and Backprop algorithms on the real data. A video showing real-time control results on the Baxter can be found \href{https://rse-lab.cs.washington.edu/se3-structured-deep-ctrl/}{here}.

\textbf{Speed: } \seposenets\ optimize errors directly in the low-dimensional pose space for control. This leads to significant speedups compared to the baselines: while both the flow and \senets\ can operate at around 10Hz (excluding the data-association pipeline), \seposenets\ run in real-time (30Hz) including the pose detection part.

\vspace{-1mm}
\section{Discussion} \label{sec:future}

This paper presents \seposenets, a framework for learning predictive models that enable control of objects in a scene. In the context of a robot manipulator, we showed they solve this problem by learning a predictive model for the individual parts of the manipulator, as in prior work \cite{byravan2017se3}. Additionally, \seposenets\ learn a consistent \emph{pose space} for these parts, essentially learning to \emph{detect} the 6D poses of manipulator parts in the raw depth images.  This detection capability enables \seposenets\ to solve the data association problem that is crucial for relating the current observation of the manipulator to a desired target observation. The difference between these poses can be used to generate control signals to move the manipulator to its target pose, similar to visual servoing applied to an image of the manipulator. We also showed how the learned network can be used to determine the gradients needed for the control signals. Our experiments show that \seposenets\ generate control superior to representations learned by previous techniques, even when these are provided with external data associations. Furthermore, in addition to providing data associations, \seposenets\ allow us to compute controls directly in the low dimensional pose space, enabling far more efficient control than techniques that operate in the raw perception space. Crucially, all these abilities are learned in a single framework based on raw data traces solely annotated with frame-to-frame point cloud correspondences. 

Overall, the control performance shown by our \seposenets\ is extremely encouraging and provides a strong proof of concept that such networks can learn a consistent pose space that provides long-range correspondences and fast reactive control. While this provides reason to rejoice, there are a multiple areas for improvement: 1) As shown in the real robot results, \seposenets\ (and other baselines) have difficulties handling joints further down  the kinematic chain (joints 4,5,6 for the Baxter) whose motions are significantly correlated to the motions of the joints above. Additionally, the end-effector has poor visibility on depth images. Adding state information in the form of encoder data significantly alleviates this issue but does not fully solve it. There are potentially multiple ways to improve the model to tackle this problem, including curriculum and active learning along with better regularization and physical grounding of the pose space to remove inconsistencies. 2) A key area for future work is in extending our system to interact with and manipulate external objects. Here, a consistent pose space for objects in the scene will enable the robot to plan its motion toward the objects, enabling smooth interactions. 3) Finally, while we have shown that \seposenets\ can be used for single-step reactive control, we would like to do long-term planning using model based techniques such as iterative LQG \cite{todorov2005generalized} to leverage the full strength the latent pose space, i.e., fast real-time rollouts directly in the pose space.


\section*{Acknowledgments}
\vspace{-1mm}
This work was funded in part by the National Science Foundation under contract number NSF-NRI-1637479 and STTR number 1622958 awarded to LULA robotics. We would also like to thank NVIDIA for generously providing a DGX used for this research via the UW  NVIDIA AI Lab (NVAIL).

\vspace{-1.5mm}
{\small
\bibliographystyle{IEEEtran}
\bibliography{references}
}

\end{document}